\documentclass[preprint,12pt]{elsarticle}

\usepackage{amssymb}

\newcommand{\namedcomment}[3]{%
  \newcommand{#1}[1]%
  {\todo[inline,linecolor=#3,%
    backgroundcolor=#3!25,%
    bordercolor=#3]{\textbf{#2:} ##1}}%
}
\namedcomment{\stefano}{ST}{green}
\namedcomment{\michele}{MI}{red}
\namedcomment{\andrea}{AN}{orange}
\namedcomment{\vin}{VI}{purple}
\usepackage{soul}
\usepackage{todonotes}
\usepackage{placeins}
\usepackage{enumitem}
\usepackage{caption}
\usepackage{hyperref}
\usepackage{graphicx}
\usepackage{url}
\usepackage{amsmath}
\usepackage{subcaption}
\graphicspath{ {immagini/} }
\usepackage{booktabs}
\usepackage{multirow}
\usepackage{marginnote}
\usepackage[utf8]{inputenc}
\usepackage{adjustbox}
\usepackage[ruled,vlined,linesnumbered]{algorithm2e}

\usepackage{mathtools}
\usepackage{layouts}
\usepackage{amsfonts}
\usepackage{epstopdf}
\usepackage{xspace}
\usepackage{tabularx}

\newcommand{\modelname}{STREED-Net\xspace}
\usepackage[stable]{footmisc}

\usepackage[nolist]{acronym}

\journal{Information Sciences}

\begin{document}

\begin{frontmatter}



\title{Listening to the city, attentively: A Spatio-Temporal Attention Boosted Autoencoder for the Short-Term Flow Prediction Problem}

\author[label1]{Stefano Fiorini}
\author[label1]{Michele Ciavotta}
\author[label1]{Andrea Maurino}
\affiliation[label1]{organization={University of Milano-Bicocca, Department of Informatics, Systems and Communication},
            addressline={Viale Sarca, 336},
            city={Milan},
            postcode={20126},
            country={Italy}}



\begin{abstract}
In recent years, studying and predicting alternative mobility (e.g., sharing services) patterns in urban environments has become increasingly important as accurate and timely information on current and future vehicle flows can successfully increase the quality and availability of transportation services. This need is aggravated during the current pandemic crisis, which pushes policymakers and private citizens to seek social-distancing compliant urban mobility services, such as electric bikes and scooter sharing offerings. However, predicting the number of incoming and outgoing vehicles for different city areas is challenging due to the nonlinear spatial and temporal dependencies typical of urban mobility patterns. In this work, we propose STREED-Net, a novel deep learning network with a multi-attention (spatial and temporal) mechanism that effectively captures and exploits complex spatial and temporal patterns in mobility data. The results of a thorough experimental analysis using real-life data are reported,  indicating that the proposed model improves the state-of-the-art for this task.
\end{abstract}




\end{frontmatter}


\begin{acronym}
\acro{ANN}{Artifical Neural Network}
\acro{CMU}{Cascade Multiplicative Unit}
\acro{CNN}{Convolutional Neural Network}
\acro{GRU}{Gated Recurrent Unit}
\acro{HA}{Historical Average}
\acro{LSTM}{Long Short-Term Memory}
\acro{MU}{Multiplicative Unit}
\acro{ReLU}{Rectified Linear Unit}
\acro{RNN}{Recurrent Neural Network}
\acro{SVR}{Support Vector Regression}
\acro{BN}{Batch Normalization}
\acro{DNN}{Deep Neural Network}
\end{acronym}

\section{Introduction}\label{sec:introduction}
In recent years, many researchers have studied and created models to describe and predict mobility dynamics, or flow prediction in urban areas. 
This interest is motivated by the need to comprehend displacement dynamics, which are also rapidly changing due to alternative electric and shared public transport systems, to define effective regulatory strategies for human mobility and freight transport in the smart city~\cite{zheng2014urban}. 
A clear example of the public decision-maker's regulatory action concerns the capacity constraints on public transport (in Europe) imposed to control the spread of the SARS-CoV-2 virus that significantly reshaped mobility habits in large urban areas. 

Mobility models are traditionally deployed to plan actions over the long term, but acting in a timely (even preventive) manner is becoming increasingly necessary to achieve a high quality of service and availability. This need has grown in recent years, mainly due to the introduction of new sharing services, which according to the latest Moovit report \cite{Moovit}, are increasingly appreciated by consumers for the shorter commuting time and the greater flexibility to reach areas not well served by public transport. 

In this regard, this research draws on the following two guiding scenarios. In the first one, we consider a company, which runs shared mobility services, that exploits demand forecasting models to improve short-term resource planning. 
In the second scenario, a public administration wants to predict the number of vehicles entering the different areas of the city to identify in advance possible traffic jam conditions. 
In both cases, the area of interest is divided into a grid, each element of which is a region. The number of vehicles entering (Inflow) and exiting (Outflow) each region must be predicted.
This problem has inherently spatio-temporal characteristics; evidently, the vehicular flow entering (exiting) a region does not only present temporal dependencies (time of day, flow in the previous hours) but also spatial dependencies as it strongly depends on the traffic leaving (entering) adjacent areas. 
Formally, such considerations relate to two widely recognized properties in the study of displacement dynamics~\cite{Liu2019}, namely temporal and spatial correlation. 
Mobility data are innately continuous time series generally not associated with abrupt changes. 
This means that displacement dynamics in temporally close periods share similarities, and this phenomenon is all the more true when the data sampling frequency increases. 
Moreover, different neighboring areas featuring similar functional characteristics (e.g., residential, commercial and industrial areas) often show correlated traffic patterns. 
However, the models proposed in the literature for predicting vehicular flows entering or leaving a particular area generally tend to consider all areas adjacent to the one considered as equivalently predictive. In practice, the district structure of the city tends to be irregular, which calls for a mechanism that can identify these zones and exploit this information to improve the forecast.

Finally, external factors have also a profound impact on the use of vehicles. For instance, it is well known in the literature that weather conditions and the days of the week (workdays vs. weekend) affect displacement dynamics, especially for lightweight transport means like bikes.


These considerations have steered the design of our proposal, STREED-Net (\textbf{S}patio \textbf{T}emporal \textbf{RE}sidual \textbf{E}ncoder-\textbf{D}ecoder \textbf{N}etwork) novel effective flow prediction deep learning network.

The main contributions of this paper can be summarized as follows:

\begin{enumerate}
    \item We propose a novel prediction architecture that includes two different attention blocks to acquire customized temporal and spatial information, i.e. able to adapt specifically to the city and the means of transport considered, identifying districts in the city.
    \item Moreover, to the best of our knowledge, \modelname is the first autoencoder architecture that combines the use of convolutional blocks with residual connections, a series of \ac{CMU} and two different attention mechanisms.
    \item Finally, this work presents a methodologically sound comparative performance assessment of various models from the literature on real-life datasets. The analyses presented consider different types of loss functions and KPIs. Results indicate that \modelname outperforms the considered state-of-the-art approaches.
\end{enumerate}

The rest of this paper is organized as follows. 
In \autoref{sec:sota}, the literature on techniques used in flow and traffic prediction is analyzed.
Section \ref{sec:problem} defines the flow prediction problem in urban areas, while in \autoref{sec:framework}, the core deep learning techniques exploited in this work and the proposed framework are described in detail.
In section~\ref{sec:experiments} data and results of experiments are presented and analyzed. 
Finally, the conclusions and recommendations for future work are discussed in \autoref{sec:conclusions}.

\section{Related Work}\label{sec:sota}


Several studies have addressed the problem of predicting vehicle flows in urban environments.
This problem has been initially modelled as a time series prediction problem for each city area and approached through classical statistical methods at first, and \acp{ANN} (e.g., deep learning) later.
In particular, different statistical methods have been applied, including autoregressive integrated moving average (ARIMA)~\cite{b11}, Kalman filtering \cite{b12}, and their variants, as well as other classical approaches such as Bayesian networks \cite{b13}, Markov chain \cite{b14}, and \ac{SVR} models \cite{b15}.
Others approaches have used k-means clustering, principal component analysis, and self-organizing maps to mine spatio-temporal performance trends \cite{b10}. 
However, classical statistics models shows some weaknesses when applied to the flow prediction problem, namely they are unable to capture the spatial dependencies between the various areas because data for each region of the city are considered as independent time series, and ii) they fail to capture the nonlinear relationship between space and time, which is essential for reliable prediction.
Further studies overcame these downsides by considering spatial relationships~\cite{10.1145/3097983.3098018} and external factors (e.g., environment and weather conditions~ \cite{QIAN201531}) within traditional time-series prediction methods.

\acp{ANN} have exploited in flow predictions for their capability of capturing the non-linear spatial and temporal relationships within data. 
Initial works using \acp{ANN} followed two main approaches. The first one exploits variants of \acp{RNN}~\cite{b17} such as \textit{i}) \ac{LSTM}~\cite{b16} and \textit{ii}) \ac{GRU}~\cite{cho2014learning}, whose architectures can effectively capture both the long-term pattern and short-term fluctuation of time series.
The second research line applies models based on \acp{CNN} to identify spatial dependencies in traffic networks, treating dynamic traffic data as a sequence of frames~\cite{b18}. 
However, these neural networks in their standard configuration (derived from the image recognition field) can only identify either spatial or temporal patterns of the traffic flow data, respectively.

Spatial and temporal information are inherent in traffic data, making it essential to consider both aspects at the same time when predicting mobility dynamics.
In this direction, deep learning-based approaches have been recently proposed, which exploits architectures able to capture spatial and temporal patterns, including 2D convolutions and residual units~\cite{Zhang2018, b23}, 3D convolutions \cite{b22}, 3D convolutions and \ac{LSTM}~\cite{3D-CLoST}, a combination of 2D and 3D convolutions \cite{guo2019deep}, autoencoder architecture with Multiplicative Cascade Unit (CMU) \cite{xu2018predcnn}.
In recent years, with the development of graph convolutional networks \cite{kipf2016semi}, which can be used to capture the structural characteristics of the graph network, we are witnessing their use in the field of traffic prediction. In \cite{li2017diffusion} the authors propose DCRNN, it is a model that captures the characteristic space through random walks on the graphs, and the temporal feature through the encoder-decoder architecture, while in \cite{zhao2019t} they apply the temporal graph convolutional network (T-GCN) model, which is in combination with the graph convolutional network (GCN) and gated recurrent unit (GRU). Finally, in \cite{PENG2021401} the authors propose a method of forecasting the traffic flow based on dynamic graphs: the traffic network is modeled by dynamic probability graphs. The convolution of the graph is performed on the dynamic graphs to learn the spatial features, which are then combined with the LSTM units to learn the temporal features.
We provide a more detailed description of a selection of the approaches mentioned above in \autoref{sec:experiments}. 
%
%
\section{Problem Statement}\label{sec:problem}

Given a tessellation of the area of interest (henceforth referred to as \textit{city}) in regularly-shaped regions, a set of historical observations regarding trajectories of vehicles within the city and, possibly, other spatial and non-spatial data sources for a reference time horizon $T_H$ of $H$ time points, the \textit{citywide vehicle flow prediction problem}~\cite{b22} is defined as the problem of minimizing the prediction error for vehicle \textit{Inflow} and \textit{Outflow} at time $t^\prime$ that is the first time point after $T_H$. 


In the literature, there are several definitions of location/region with different granularity and different semantic meaning~\cite{b31}. However, when it comes of traffic forecasts, the majority of works uses a rectangular tessellation, which maximizes the number of neighboring areas.   
Similarly, in this study, the geographical space of interest (city) is logically partitioned into a regular grid of size $N \times M$ oriented by longitude and latitude~\cite{Zhang2018}. Each element of the grid is termed \textit{region} and is  addressable through a pair of coordinates $(n,m)$ corresponding to the $n^{th}$ row and the $m^{th}$ column of the grid.

\begin{figure}[t]
\begin{subfigure}[t]{0.5\textwidth}
\centering
\includegraphics[width=.65\textwidth]{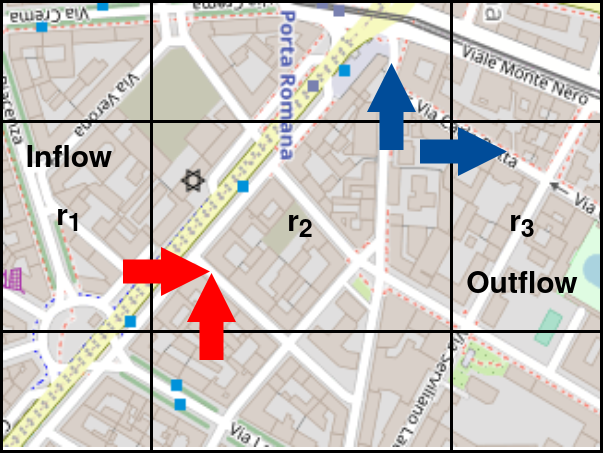}
\caption{Inflow and Outflow}
\label{fig: Inflow and Outflow}
\end{subfigure}
~
\begin{subfigure}[t]{0.5\textwidth}
\centering
\includegraphics[width=.9\textwidth]{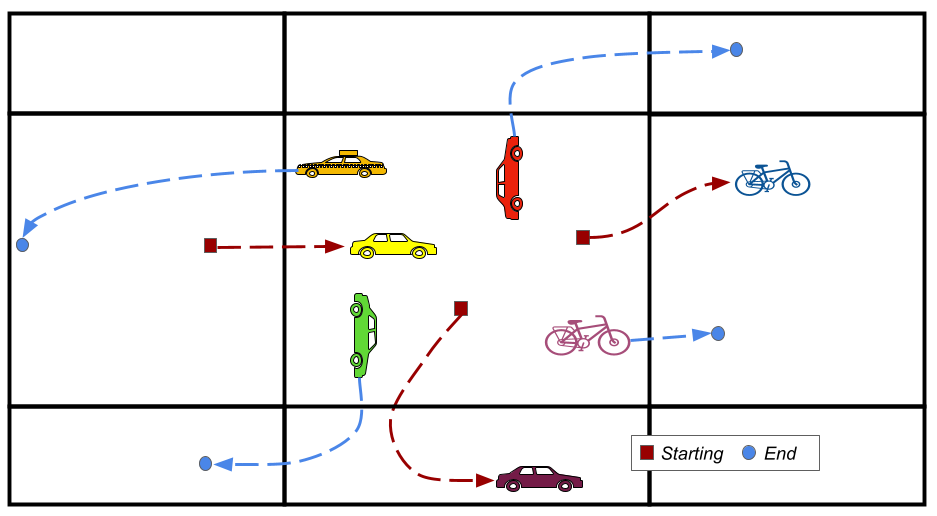}
\caption{Measurement of flows}
\label{fig: Measurement of flows}
\end{subfigure}
\end{figure}

The term Inflow (Outflow, respectively) refers to the number of vehicles entering (leaving) a specific region in the considered time unit (\autoref{fig: Inflow and Outflow})~\cite{b22}. 
More specifically, the Inflow (Outflow) indicates the number of pedestrians, cars, public transport and sharing vehicles entering (leaving) the region in a certain time period. 
As shown in \autoref{fig: Measurement of flows}, by analyzing the movement data of the vehicles, it is possible to obtain the Inflow and Outflow matrix, which encompass the information about displacements between the areas of the city at each time $t$.
More in detail, let $\tau_i = \{s^1_i, s^2_i,\ldots s^t_i \}$ be a trajectory where $s_i^t$ represents the position of vehicle $i$ at time $t$, and let $\mathcal{T}$ be a collection of trajectories. The Inflow (Outflow, respectively) of a region $(n,m)$ at time $t$, namely $\iota^{t}_{n,m}$ ($\omega^{t}_{n,m}$)  can be formally defined as in equation~\eqref{eq:1} (\ref{eq:2}, respectively).

\begin{equation} \label{eq:1}
\footnotesize
    \iota^{t}_{n,m} = \sum_{\tau_i \in \mathcal{T}} \phi_{\iota} (\tau_i, t, n,m)
\end{equation}
\begin{equation} \label{eq:2}
\footnotesize
    \omega^{t}_{n,m} = \sum_{\tau_i \in \mathcal{T}} \phi_{\omega}(\tau_i, t, n,m)
\end{equation}
where 
\begin{equation*}
  \footnotesize
\phi_{\iota}(\tau_i,t,n,m)=
\begin{cases}
1, \quad \text{if}\; s_i^{t-1} \notin (n,m) \wedge s^{t}_i \in (n,m)  \\
0, \quad\text{otherwise.}
\end{cases}  
\end{equation*}
and
\begin{equation*}
\footnotesize
\phi_{\omega}(\tau_i,t,n,m)=
\begin{cases}
1, \quad\text{if}\; s_i^{t-1} \in (n,m) \wedge s^{t}_i \notin (n,m)  \\
0, \quad\text{otherwise.}
\end{cases}  
\end{equation*}

Finally, the state of the vehicular flow at time $t$ can be represented by a tensor (also referred to as \textit{frame} in what follows) $F_{t} \in \mathbb{R}^{N\times M\times C}$, where $C$ indicates the number of flow variables considered in the analysis, in this specific case $C = 2$ (Inflow/Outflow), whereas $N\times M$ is the total number of regions in the city.
Then, to take into account the temporal dependence, over the time horizon $T$ (divided into $H$ time points), the flow representation is extended to a tensor of four dimensions $F \in \mathbb{R}^{H\times N\times M\times C}$, which represents the main input to our problem.
The problem at issue then becomes predicting $F_t$ given a \textit{volume}, that is a sequence of past tensors $\mathcal{V}\subset F$. 
It is worth noting that the resulting problem shows several similarities with the frame prediction problem~\cite{3D-CLoST} since the tensor $F$ can be seen as a four dimensional volume composed of $H$ consecutive images, each of which featuring $C$ channels. 
\section{\modelname}\label{sec:framework}
\modelname is a \textit{Autoencoder}-based deep learning model that combines convolutions and \acp{CMU} with two different types of \textit{Attentions} (spatial and temporal). 
This section presents \modelname, detailing its components and relationships, prefacing it with a brief introduction to the main underpinning concepts, namely the autoencoder architecture and the attention mechanism. 

\subsection{Prerequisites}
\textbf{Autoencoder architecture.} Given a set of unlabeled training examples $\{x^1,x^2,x^3, ...\}$, where $x^i \in \mathbb{R}^n$, an autoencoder neural network is an unsupervised learning algorithm that applies backpropagation setting the target values to be equal to the inputs $y(i)=x(i)$. It is a neural network that is trained to learn a function $h_{W,b}(x) = \hat{x} \approx x$, where $W$ and $b$ are weights and biases of the \ac{ANN}, respectively.
In other words, an autoencoder is a learned approximation of the identity function, so as to output $\hat{x}$ that is as much as possible similar to $x$. 
The overall network can be decomposed into two parts: an encoder function $h = f(x)$, which maps the input vector space onto an internal representation, and a decoder that transforms it back, that is $\hat{x} = g(h)$.
This type of architecture has been applied successfully to different difficult tasks, including traffic prediction \cite{xu2018predcnn}.

\textbf{Attention mechanism.} 
In \acp{DNN} Attention Mechanism helps focus on important features of the input, shadowing the others. This paradigm is inspired by the human neurovisual system, which quickly scans images and identifies sub-areas of interest, optimizing the usage of the limited attention resources~\cite{ungerleider2000mechanisms}. Similarly, the attention mechanism in \acp{DNN} determines and stresses on the most informative features in the input data that are likely to be most valuable to the current activity.

Recently, attention has been widely applied to different areas of deep learning, such as natural language processing \cite{bahdanau2014neural}, 
image recognition \cite{zheng2017learning}, image captioning \cite{xu2015show}, image generation \cite{gregor2015draw} and traffic prediction \cite{8880638}.

\begin{figure*}[htb]
\centering
\includegraphics[width=\textwidth, height= 4.5cm]{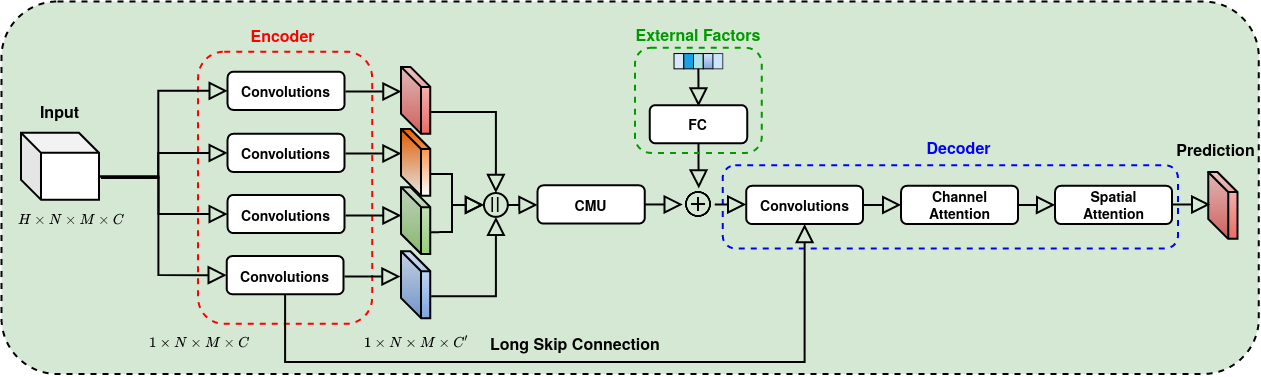}
\caption{\modelname Architecture}
\label{fig: Model}
\end{figure*}


\subsection {Encoder}
\begin{figure}[htb]
\centering
\includegraphics[width=0.7\textwidth]{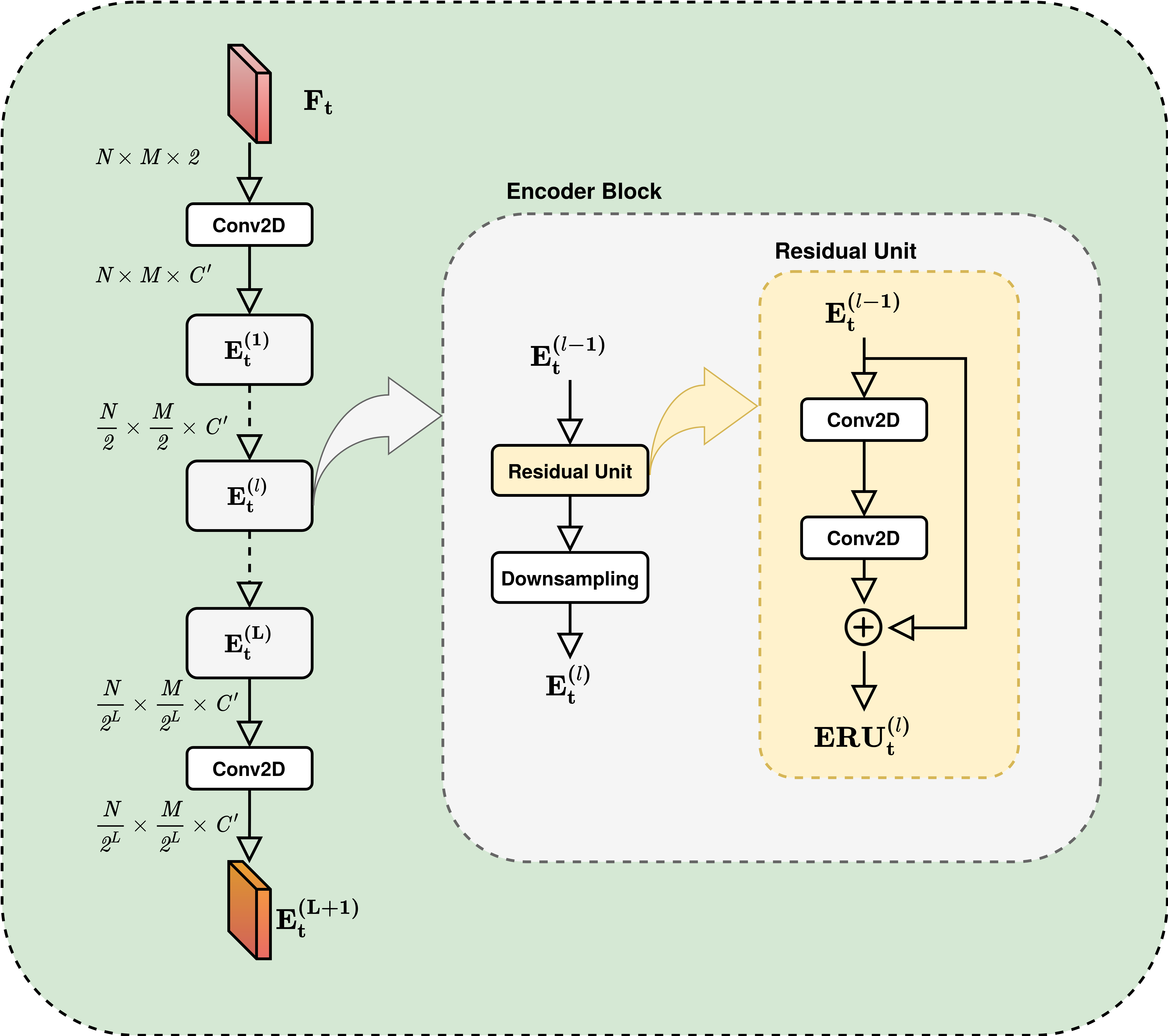}
\caption{Encoder}
\label{fig: Encoder}
\end{figure}

The encoder structure depicted in \autoref{fig: Encoder} is the first block of the \modelname architecture. 
It is composed of an initial convolutional layer, a series of \textit{residual units}, and a final convolution layer. 
Unlike similar approaches (e.g., STAR~\cite{b23}), the proposed encoder structure introduces three novel aspects:
\textit{i)} each layer is time-distributed, meaning that the model learns from a sequence of frames (for time coherence) instead of focusing on each frame singularly.
\textit{ii)} it applies further convolutions after the residual unit, so that to reduce the frame size and \textit{iii)} it applies a \acfi{BN} after each convolution 
to avoid gradient disappear/explode problems and achieve faster and more efficient reported optimization \cite{ioffe2015batch, NEURIPS2018_905056c1}.

Unlike other works from the literature, where the distant temporal information is also used (from the previous day and previous week), 
the encoder takes as input a four-dimensional tensor $F \in \mathbb{R}^{H\times N\times M\times 2}$.
This tensor is a sequence of consecutive three-dimensional frames conveying flow information of nearby periods (with regard to the prediction time $t^\prime$). 
Such a tensor (also referred to as \textit{closeness }in the literature~\cite{b22}) is obtained by selecting $p$ points preceding the prediction time $t^\prime$, i.e., the sequence $ [F_ {t^\prime-p}, F_ {t^\prime- (p +1)}, ...., F_ {t^\prime-1}] $. In this way, \modelname can focus on the most recent dynamics only. 
Each frame in $F$ is processed by the convolution layer to extrapolate spatial information.
It is worth noting that in Figure~\ref{fig: Model}, the encoder is represented by a collection of identical blocks in parallel execution on the input frames instead of (as in reality) a single convolution applied sequentially. 
Such a representation is used to highlight that a time-distributed layer is trained by taking into account all input frames simultaneously. 
The use of this approach leads the model to identify temporal (that is, inter-frame) dynamics, rather than looking only to spatial dependencies within each frame. 

Each convolutional layer is followed by a \acfi{ReLU} activation function and a \textit{BN} layer. 
Formally, we have:
\begin{equation}
    E_t^{(0)} = \textit{BN}(\textit{ReLU}(W_e^{(0)} \ast F_t + b_e^{(0)}))
\end{equation}
where $E_t^{(0)}$ corresponds to the output of the first convolutional layer, $F_t$ is one of $p$ frames in input to the model and  $\ast$ the convolution operator. $W_e^{(0)}$ and $b_e^{(0)}$ are the weights and biases of the respective convolutional operation.  
Next, $\mathbf{L}$ encoder blocks (see Figure~\ref{fig: Encoder}) are placed. Each of these blocks is composed of a residual unit followed by a downsampling layer: 
\begin{equation}
E_t^{(l)} = \textit{Downsampling}(\textit{ResUnit}(E_t^{(l-1)}))
\end{equation}
where $E_t^{(l-1)}$ and $E_t^{(l)}$ correspond respectively to the input and output of the encoder block; $l$ takes values in $\{1, ..., \mathbf{L}\}$. The residual units have been implemented as a sequence of two convolutional layers whose output is eventually summed to the block input. 
Mathematically, in \modelname the residual unit are defined as follows:
\begin{align}
&c_1 =\textit{BN}(\textit{ReLU}(W^{(l)}_1 \ast E_t^{(l-1)} + b_1^{(l)})) \\
&c_2 =\textit{BN}(\textit{ReLU}(W_2^{(l)} \ast c_1 + b_2^{(l)})) \\
&ERU_t^{(l)} = E_t^{(l-1)} + c_2
\end{align}
where $E_t^{(l-1)}$ is the residual unit input and $ERU_t^{(l)}$ (\textit {Encoder Residual Unit}) is used to indicate the result of $\textit{ResUnit}(E_t^{(l-1)})$. $\ast$ is the convolution operator, $W_1^{(l)}$, $W_2^{(l)}$ and $b_1^{(l)}$, $b_2^{(l)}$ are the weights and biases of the respective convolutional operations. 

For what concerns the \textit{Downsampling}, it has been implemented as: 
\begin{equation}
    E_t^{(l)} = \textit{BN}(\textit{ReLU}(W_{ds}^{(l)}\ast ERU_t^{(l)}+b_{ds}^{(l)})
\end{equation}
where $W_{ds}^{(l)}$, $b_{ds}^{(l)}$ and $\ast$ indicate a convolutional layer with \textit{kernel size} and \textit{stride} parameters set to halve the height and width of the input frame. 

The rationale behind the design of this architecture is threefold: \textit{i)} a deep structure is needed for the model to grasp dependencies not only among neighboring regions but also among distant areas; \textit{ii)} \textit{Deep} networks are difficult to train as they present both the problem of the explosion or disappearance of the gradient and a greater tendency to overfitting due to the large number of parameters. To try to avoid these obstacles and to make the training model more efficient, we introduced residual units. 
Finally, \textit{iii)} the downsampling layers were introduced to ensure translational equivariance \cite{goodfellow2016deep}.

Finally, the encoder structure ends with a closing convolution-\ac{ReLU}-\ac{BN} sequence, which has as its main objective to reduce the number of \textit{feature maps}.
In this way, the next architectural component (i.e., the Cascading Hierarchical Block) will receive and process a smaller input, reducing the computational cost of the \ac{CMU} array. 
The encoder output is:
\begin{equation}
   E_t^{(L+1)} = \textit{BN}(\textit{ReLU}(W_e^{(L)}\ast E_t^{(L)}+b_e^{(L)}))
\end{equation}

The output of the encoder is a tensor $E^ {(L + 1)} \in \mathbb {R} ^ {H \times N / 2 ^ \mathbf{L} \times M / 2 ^ \mathbf{L} \times C^\prime}$, 
where $C^\prime$ is the number of \textit{feature maps} generated by the last convolution of the encoder. 

\subsection{Cascading Hierarchical Block}

A connection section between the encoder and the decoder is provided to handle the temporal relationships among the frames.
Unlike what is proposed in other works that combine the use of \ac{CNN} with the use of \ac{RNN} such as \ac{LSTM} \cite{Ranjan2020}, \modelname implements a Cascading Hierarchical Block with \aclp{CMU} (CMU)~\cite{xu2018predcnn},
which computes the hidden representation of the current state directly using the input frames of both previous and current time steps, rather than what happens in recurrent networks that model the temporal dependency by a transition from the previous state to the current state. 
This solution is designed to explicitly model the dependency between different time points by conditioning the current state on the previous state, improving the model accuracy; incidentally, it also reduces training times.

The fundamental constituent of \ac{CMU} architecture is the \acfi{MU} \cite{kalchbrenner2017video}, which is a non-recurrent convolutional structure whose neuron connectivity, except for the lack of residual connections, is quite similar to that of \ac{LSTM}~\cite{b9}; the output, however, only depends on the single input frame $h$. 
Formally, \ac{MU} is defined by the following equation set:
\begin{align}
    &g_1 = \sigma(W_1 * h + b_1) \\
    &g_2 = \sigma(W_2 * h + b_2) \\
    &g_3 = \sigma(W_3 * h + b_3) \\
    &u = tanh(W_4 * h + b_4) \\
    &MU(h; W) = g_1 \odot tanh(g_2 \odot h + g_3 \odot u)
\end{align}
where $\sigma$ is the sigmoid activation function, $*$ the convolution operator and $\odot$ the element-wise multiplication operator. $W_1 \sim W_4$ and $b_1 \sim b_4$ are the weights and biases of the respective convolutional gates and $W$ denotes all \ac{MU} parameters.

\begin{figure}[tbh]
\centering
\includegraphics[width=.8\textwidth]{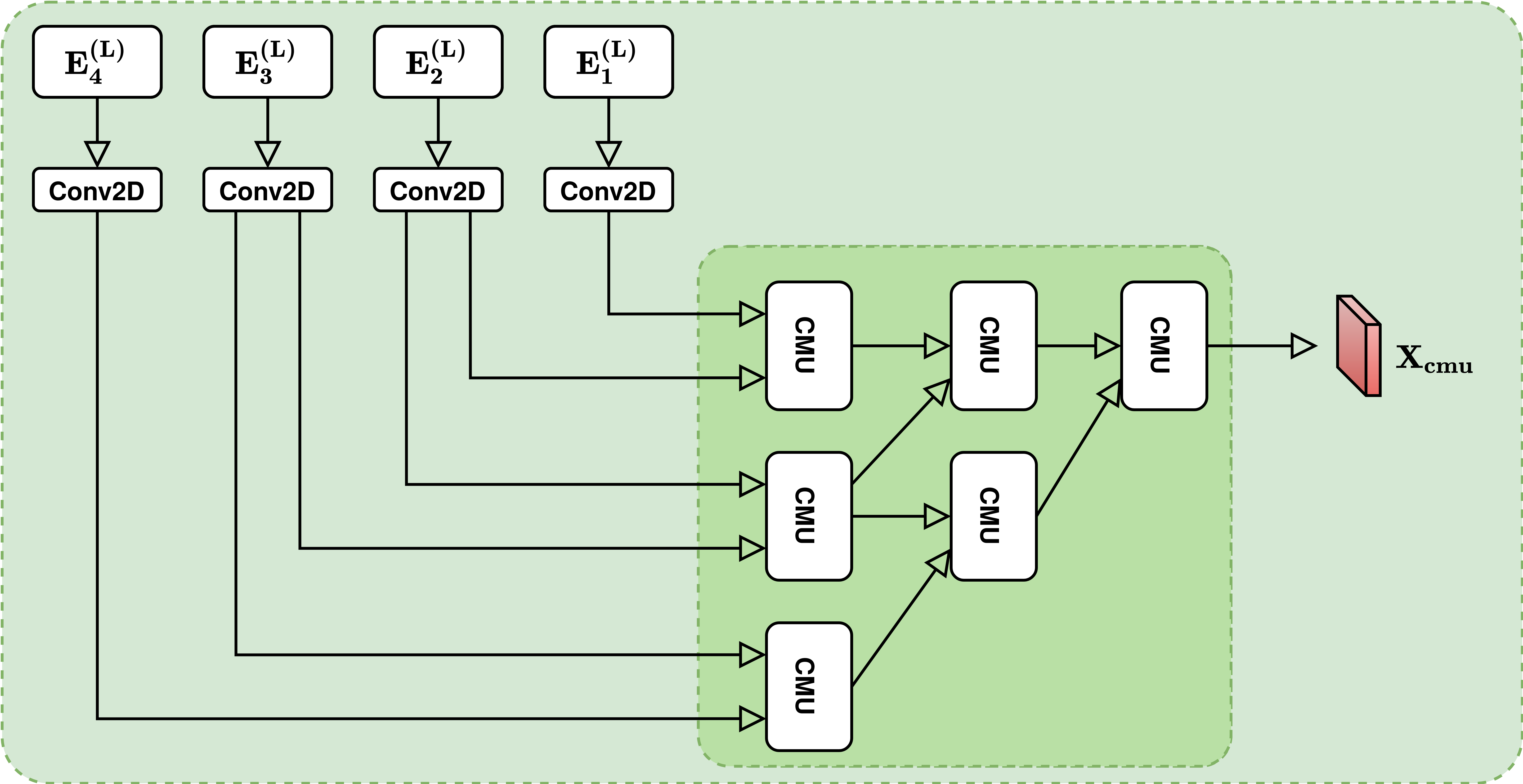}
\caption{Cascading Hierarchical Block}
\label{fig: Cascading Hierarchical Block}
\end{figure}

\ac{CMU} incorporates three \acp{MU}. Unlike \ac{MU}, \ac{CMU} accepts two consecutive frames as input to model explicitly the temporal dependencies between them. 
The more recent frame in time is inputted to a \ac{MU} to capture the spatial information of the current representation. 
The older frame is instead processed by two \acp{MU} in sequence to overcome the time gap. The partial outputs are then added together and finally, thanks to two gated structures containing convolutions along with non-linear activation functions, the output of the \ac{CMU} ($X_t^{l+1}$) is generated. 
\ac{CMU} is described by the following equations:
 \begin{align}
    &h_1 = MU(MU(E^l_{(t-1)}; W_1); W_1) \\
    &h_2 = MU(E_t^l; W_2) \\
    &h = h_1 + h_2 \\
    &o = \sigma(W_o * h + b_o) \\
    &X_t^{l+1} = o \odot tanh(W_h * h + b_h)
\end{align}
where $W_1$ and $W_2$ are the parameters of the \ac{MU} in the left branch and of the \ac{MU} in the right branch respectively, $W_o$, $W_h$, $b_o$ and $b_h$ are the weights and biases of the corresponding convolutional gates. 
The cascading hierarchical block uses CMUs to process all frames at the same time (see Figure~\ref{fig: Cascading Hierarchical Block}):
 \begin{align}
    &X_{cmu} = CascadeCMU(E^{(L+1)})
\end{align}
where $X_{cmu} \in \mathbb{R} ^{N\times M\times C^\prime}$.

\subsection{External Factors}

To integrate external information, such as the day of the week, holidays, and weather conditions \modelname features a specific input branch. 
The input is a one-dimensional vector that contains information that refers to prediction time $t^\prime$. 

Through the use of two fully connected overlapping layers, this information is conveyed, encoded, into the mainstream of the network.
The first level is used to embed each sub-factor, while the second reshapes the external factors embedding space to match the size of the \textit{\ac{CMU} output vector}. 

\subsection{Decoder} \label{sec: decoder}

The decoder is the last component of \modelname and its task is to generate the flow prediction starting from the latent representation that corresponds to the output of the cascading hierarchical block.

\begin{figure}[htb]
\centering
\includegraphics[width=0.7\textwidth]{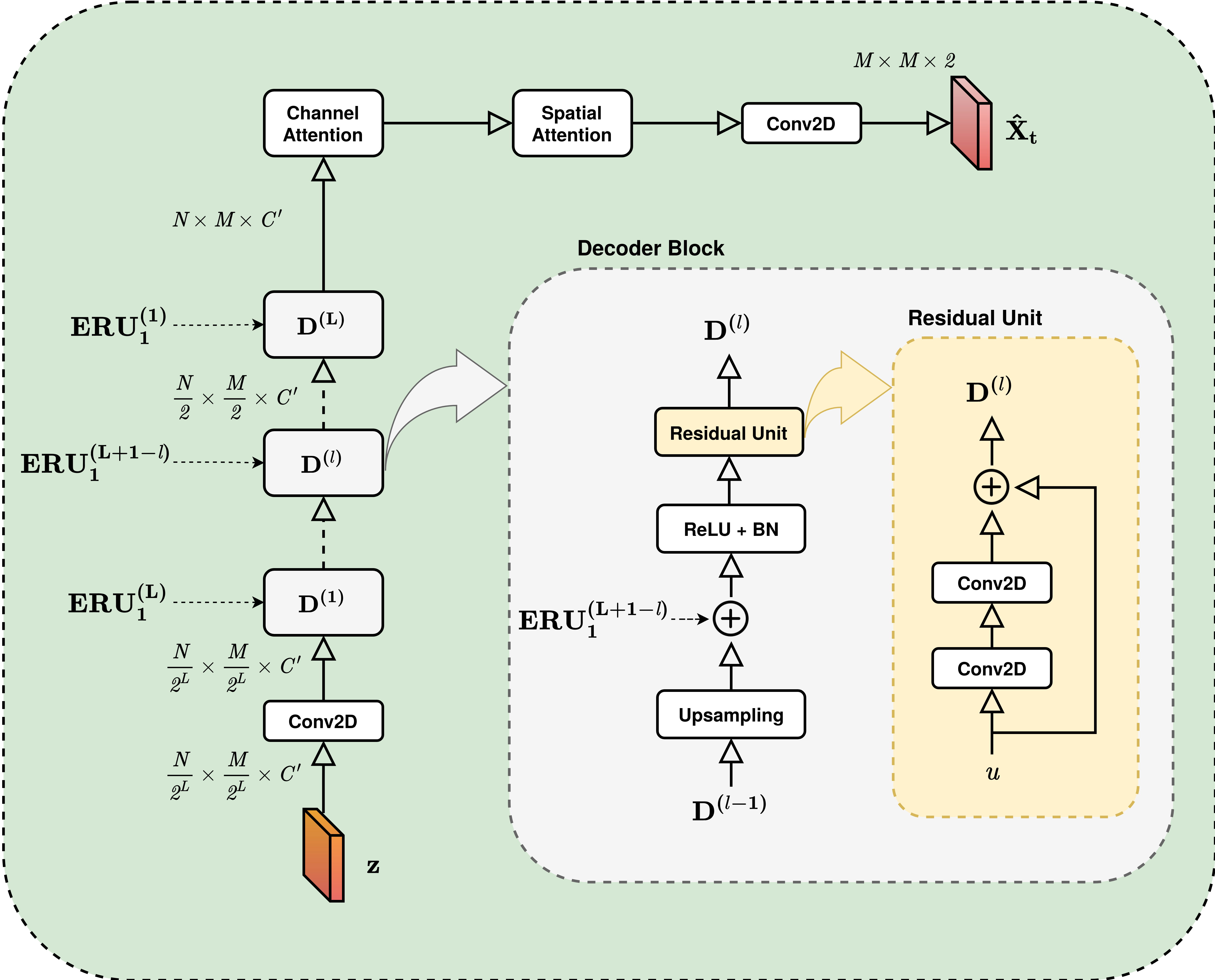}
\caption{Decoder}
\label{fig: Decoder}
\end{figure}

As shown in \autoref{fig: Decoder}, the decoder takes as input a tensor $z = X_{cmu} + X_{ext}$, where $z \in \mathbb{R} ^{N\times M\times C^\prime}$,  which is the result of the sum of the outputs of the hierarchical structure and the network dedicated to incorporate external factors. 
$X_ {ext}$ is added at this point of the network to allow the model to use the information extracted from the external factors during the reconstruction phase. 


The decoder architecture features a structure that is somehow symmetrical to that of the encoder with an array of residual units preceded and followed by a convolutional layer (\autoref{eq:dec_first}). 

\begin{equation}\label{eq:dec_first}
\small
    D^{(0)} = \textit{BN}(\textit{ReLU}(W_{d}^{(0)} * z + b_{d}^{(0)}))
\end{equation}

Nevertheless, this symmetry is breached by the presence of two significant differences. 
The first one is the presence of a long skip connection before every residual unit. The long skip connection is used to improve the accuracy and to recover the fine-grained details from the encoder.
Another benefit is a significant speed up in model convergence~\cite{DBLP:journals/corr/SzegedyIV16}.
A generic decoding block $D^{(l)},\;\forall l\in\{1\dots L\}$ can be formally defined as the sequential application of the following three operations:
\begin{align}
\small
&sc^{(l)} =\textit{Conv2DTranspose}(D^{(l-1)})+ERU^{(L+1-l)}_1\\
&U^{(l)} =\textit{BN}(\textit{ReLU}(sc^{(l)})\\
&D^{(l)} = \textit{ResUnit}(U^{(l)})
\end{align}

where $D^{(l-1)} $ corresponds to the input block, \textit{Conv2DTranspose} indicates the transposed convolution operation (also known as deconvolution), which doubles the height and width of the input, and $sc^{(l)}$ (skip connection) is the sum of \textit{u} with $ERU_1^{(L+1-l)}$, i.e., the output of the remaining encoder unit at level $L+1-l$ for the most recent frame. The residual units of the decoder are structured exactly like those of the encoder.


The second difference is the presence of two attention blocks (viz. Channel and Temporal Attention) before the final convolution layer. More details are provided in the following subsections.

\subsubsection*{Channel Attention}

After the convolutional stage of the decoder, a three-dimensional tensor, referred to as $D^{(L)} \in  \mathbb{R}^{N\times M\times C^\prime}$, is obtained with the channel size $C^\prime$. 
Since the dimension of the channel also includes the temporal aspects compressed by the cascading hierarchical block, the channel attention has been introduced to identify and emphasize the most valuable channels.

\begin{figure}[htb]
\centering
\includegraphics[width=.5\textwidth]{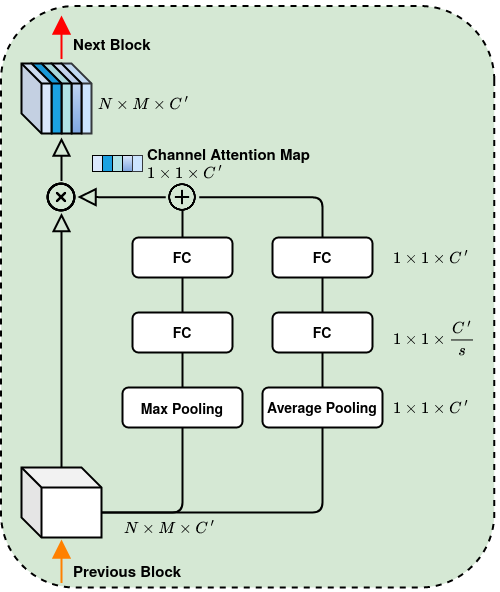}
\caption{Channel Attention Block}
\label{fig: Temporal Attention Map}
\end{figure}

\autoref{fig: Temporal Attention Map} shows the structure of the \textit{channel attention block}.
Given the input tensor, $D^{(L)}$ a channel attention map $A_c \in \mathbb{R}^{1\times 1\times C^\prime}$ is created by applying attention block deduction operations on the channels.
More in details, through the operation of global average pooling and global max pooling performed simultaneously, two different feature maps ($X^{max}$ and $X^{avg}$) of size $1\times 1\times C^\prime$ each are spawned. 
They go through two fully connected layers that allow the model to learn (and assess) the importance of each channel. 
The first layer performs a dimensionality reduction, downsizing the input feature maps to $1\times 1\times \frac{C\prime}{s}$, based on the choice of the reduction ratio $s$; the second layer restores the feature maps to their original size. 
This approach has proven to increase the model efficiency without accuracy reduction~\cite{8880638}. 
Once these two steps have been completed, the two resulting feature maps are combined into a single tensor through a weighted summation as:
\begin{equation}
    A_c = \sigma(\Lambda_1 \otimes X^{max} + \Gamma_1 \otimes X^{avg})
\end{equation}
where $\Lambda_1$ and $\Gamma_1$ are two trainable tensors with the same size as the two feature maps and $\sigma$ is the activation function. 
$\Lambda$ and $\gamma$ are set during the training phase and weight the relative importance of each element of the two feature maps.
Finally, the process of getting channel attention can be summarized as:
\begin{equation}
    D^{'} = A_c \otimes D^{(L)}
\end{equation}
where $D^{'}$ is the operation output and $\otimes$ denotes the element-wise multiplication.

\subsubsection*{Spatial attention}

\begin{figure}[htb]
\centering
\includegraphics[width=.5\textwidth]{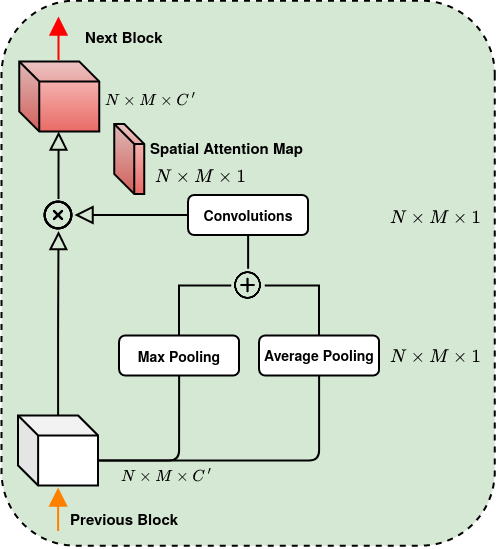}
\caption{Spatial Attention Block}
\label{fig: Spatial Attention Map}
\end{figure}

Cities are made up of a multitude of different functional areas. 
Areas have different vehicle concentrations and mobility patterns; thus, the spatial attention mechanism has the task of identifying where are located the most significant areas and scale their contribution to improve the prediction. \autoref{fig: Spatial Attention Map} presents the main internals involved in the calculation of the \textit{spatial attention Map}.

The spatial attention map $A_s \in \mathbb{R}^{N\times M\times 1}$ can be calculated by applying pooling operations along the axes of the channel to highlight informative regions~\cite{zagoruyko2016paying}.
Therefore, first the global average pooling and global max pooling operations are applied along the channel axes and, as in the channel attention block, two distinct feature maps of size $N\times M\times 1$ are obtained.
Subsequently, each single feature map passes through a convolution layer with a filter size of $4\times 4$. 
unlike what is done in other approaches from the literature \cite{woo2018cbam} where the filter is set to $7\times 7$. It is worth noting that the filter size depends on the size of the areas that make up the city. For the case studies addressed in this work (see Section~\ref{sec:experiments}), which feature rather large regions, the proposed model does not need to focus on large area clusters.

Finally, the sigmoid activation function is applied.

Also for the spatial attention block, the two feature maps are combined into a single tensor through a weighted summation:
\begin{equation}
    A_s = \sigma(\Lambda_2 \otimes X^{max} + \Gamma_2 \otimes X^{avg})
\end{equation}

Finally, the process of getting spatial attention can be summarized as:
\begin{equation}
    D^{''} = A_s \otimes D^{'}
\end{equation}


%
%
\section{Experimental Analysis}\label{sec:experiments}
This section reports on an extensive experimental evaluation of the proposed model by comparing it against several reference models (see Section~\ref{sec:other_method}) using three different performance metrics and three different case studies (detailed in Section~\ref{sec:case_studies}). 
An ablation study and the analysis of computational complexity complete the section. 

\subsection{Reference methods}\label{sec:other_method}

The proposed model is compared against the following state-of-the-art methods expressly devised to to solve the citywide vehicle flow prediction problem~\cite{b22}:

\textbf{ST-ResNet~\cite{Zhang2018}}: 
it is one of the first deep learning approaches to traffic prediction. It 
predicts the flow of crowds in and out each individual region of activity. 
ST-ResNet uses three residual networks that model the temporal aspects of \textit{proximity}, \textit{period}, and \textit{trend} separately.

\textbf{MST3D~\cite{b22}}: this model is architecturally similar to ST-ResNet. The three time dependencies and the external factors are independently modeled and dynamically merged by assigning different weights to different branches to obtain the new forecast. 
Differently from ST-ResNet, MST3D learns to identify space-time correlations using 3D convolutions.

\textbf{ST-3DNet~\cite{guo2019deep}}: the network uses two distinct branches to model the temporal components of \textit{closeness} and \textit{trend}, while the daily period is left out. Both branches start with a series of 3D convolutional layers used to capture the spatio-temporal dependencies among the input frames. In the \textit{closeness} branch, the output of the last convolutional layer is linked to a sequence of residual units to further investigate the spatial dependencies between the frames of the \textit{closeness} period.
The most innovative architectural element is the \textit{Recalibration Block}. It is a block inserted at the end of each of the two main branches to explicitly model the contribution that each region makes to the prediction.

\textbf{3D-CLoST~\cite{3D-CLoST}}: the model uses sequential 3D convolutions to capture spatio-temporal dependencies. Afterwards, a fully-connected layer encloses the information learned in a one-dimensional vector that is finally passed to an \ac{LSTM} block.
\ac{LSTM} layers in sequence allow the model to dwell on the temporal dependencies of the input. 
The output of the \ac{LSTM} section is added to the output produced by the network for external features. 
The output is multiplied by a \textit{mask}, which allows the user to introduce domain knowledge: the mask is a matrix with null values in correspondence with the regions of the city that never have Inflow or Outflow values greater than zero (such areas can exist or not depending on the conformation of the city) while it contains $1$ in all other locations.

\textbf{STAR~\cite{b23}}: this approach aims to model temporal dependencies by extracting representative frames of \textit{proximity}, \textit{period} and \textit{trend}. 
However, unlike other solutions, the structure of the model consists of a single branch: the frames selected for the prediction are  concatenated along the axis of the channels to form the main input to the network. 
In STAR as well, there is a sub-network dedicated to external factors and the output it generates is immediately added to the main network input.
Residual learning is used to train the deep network to derive the detailed outcome for the expected scenarios throughout the city.

\textbf{PredCNN~\cite{xu2018predcnn}}: this network builds on the core idea of recurring models, where previous states in the network have more transition operations than future states. 
PredCNN employs an autoencoder with \ac{CMU}, which proved to be a valid alternative to \ac{RNN}. Unlike the models discussed above, this approach considers only the temporal component of \textit{closeness} but has a relatively complex architecture. The key idea of PredCNN is to sequentially capture spatial and temporal dependencies using \ac{CMU} blocks.

\textbf{\ac{HA}}: the algorithm generates Inflow and Outflow forecasts by performing the arithmetic average of the corresponding values of the same day of the week at the same time as the instant in time to be predicted. This classical method represents a baseline in our comparative analysis as it has not been developed specifically for the flow prediction problem.

Excluding MST3D, which has been entirely reimplemented following the indications of the original paper strictly, and PredCNN, whose original code has been completed of some missing parts, for all the other models the implementation released by the original authors has been used. 
The \modelname{} code, together with all the code realized for this research work, is freely available\footnote{\url{https://github.com/UNIMIBInside/Smart-Mobility-Prediction}}.

We conclude this section by pointing out that, although the literature offers numerous proposals for deep learning models based on graphs with performances often superior to those of convolutional models, for the problem addressed in this paper, preliminary experiments that we conducted with graph-based models did not lead to satisfactory results. 
The main reason for this apparent contradiction resides in the nature of the problem considered, whose basic assumption is to be able to observe only the inflow and outflow across all areas of the city. 
Such a scenario is feasible and more realistic than one in which the trajectory or origin-destination pair of all vehicles is known but makes it impossible to create graphs with nontrivial connections (i.e., not between adjacent areas) for the problem under consideration. Graph-based models under such conditions have not proven to be sufficiently accurate for the case studies considered.


\subsection{Case studies}\label{sec:case_studies}

Three real-life case studies are considered for the experimental analysis, which differ in both the city considered (New York and Beijing) and the type of vehicle considered (bicycle and taxi). This choice allow the models to be assess on  usage patterns that are expected to be significantly distinct. Follows a brief description of the considered case studies:

\textbf{BikeNYC.} In this first case study the behavior of bicycles in New York city is analyzed. The data has been collected by the \textit{NYC Bike system} in 2014, from April 1 to September 30. Records from the last 10 days form the testing data set, while the rest is used for training. The length of each time period is of 1 hour.

\textbf{TaxiBJ.} In the second case study, a fleet of cabs and the city of Beijing are considered.  Data have been collected in 4 different time periods: July 1 2013 - October 30 2013, March 1 2014 - June 30 2014, March 1 2015 - Jun 30 2015, November 1 2015 - April 15 2016. The last four weeks are test data and the others are used for training purposes. The length of each time period is set to 30 minutes.
 
\textbf{TaxiNYC.} Finally, a data set containing data from a fleet of taxicabs in New York is considered. Data have been collected from January 1, 2009 to December 31, 2014. The last four weeks are test data and the others are used for training purposes. The length of each time period is set to one hour. This case study has been specifically created to perform a more thorough and sound experimental assessments than those presented in the literature.

The city of New York has been tessellated into $16\times 8$ regions, while the city of Beijing has been divided into $32\times 32$ areas; the discrepancy in the number of regions considered is due to the large difference in extension between the two cities. The Beijing area (16,800 km$ ^ 2 $) is 22 times bigger than the New York area (781 km$ ^ 2 $).

The Beijing taxi data set (TaxiBJ) and New York Bike data set (BikeNYC) are available via \cite{Zhang2018}; they are already structured to carry out the experiments reported in this work. As for the TaxiNYC dataset, available for experiments on GitHub\footnote{\url{https://github.com/UNIMIBInside/Smart-Mobility-Prediction/tree/master/data/TaxiNYC}}, it has been expressly built for this work by processing and structuring data available from the \textit{NYC government website}\footnote{\url{https://www1.nyc.gov/site/tlc/about/tlc-trip-record-data.page}}.

A Min-Max normalization has been applied to all data sets to convert traffic values based on the scale [-1, 1].
Note, however, that in the experiments a denormalization is applied to the expected values to be used in the evaluation.

In the three experiments, public holidays, metadata (i.e. DayOfWeek, Weekday/Weekend) and weather have been considered as external factors. Specifically, the meteorological information reports the temperature, the wind speed, and the specific atmospheric situation (viz., sun, rain and snow).

\subsection{Analysis of results}
This section presents and discusses the results of experiments performed by running \modelname{} and the models presented in Section~\ref{sec:other_method} on the three case studies.  Moreover, three different evaluation metrics are used in this study to compare the results obtained: \textit{Root Mean Square Error} (RMSE), \textit{Mean Absolute Percentage Error} (MAPE) and \textit{Absolute Percentage Error} (APE), which are defined as follows:

{\footnotesize
\begin{align}
 &R M S E=\sqrt{\frac{\displaystyle\sum_{n=1}^{N}\sum_{m=1}^M[\left(\hat{\iota}_{n,m} - \iota_{n,m}\right)^{2} + \left(\hat{\omega}_{n,m} - \omega_{n,m}\right)^{2}]}{N \times M}}
\end{align}
}%
{\footnotesize
\begin{align}
 &MAPE = 100\cdot\frac{\displaystyle\sum_{n=1}^{N}\sum_{m=1}^M\left|\frac{\left(\hat{\iota}_{n,m} - \iota_{n,m}\right)+ \left(\hat{\omega}_{n,m} - \omega_{n,m}\right)}{\left(\hat{\iota}_{n,m} - \iota_{n,m}\right)}\right|}{N \times M} \\
 &APE = 100\cdot\displaystyle\sum_{n=1}^{N}\sum_{m=1}^M\left|\frac{\left(\hat{\iota}_{n,m} - \iota_{n,m}\right)+ \left(\hat{\omega}_{n,m} - \omega_{n,m}\right)}{\left(\hat{\iota}_{n,m} - \iota_{n,m}\right)}\right| \end{align}
}%

where $\hat\iota_{n,m}$ and $\hat\omega_{n,m}$ are, respectively, the predicted Inflow and Outflow for region $(n,m)$ at time $t^\prime$ and  $N\times M$ is the total number of regions in the city.

Finally, it is worth noting that to account for and reduce the inherent stochasticity of learning-based models, each experiment was repeated ten times (replicas) using a different random seed in each replica. Mean and standard deviation are reported for each metric to provide a robust indication of the overall behavior of the compared methods.

\subsubsection{BikeNYC}
For the BikeNYC case study 
\modelname parameters have been set as follow. The number $n$ of input frames has been set to 4, the number $L$ of encoding and decoding blocks has been set to 2. This decision has been dictated by the size of the grid ($16 \times 8$): 
setting $L$ greater than 2 (for example 3) would result in an encoder output tensor of size $4\times 2\times 1\times C$
, which would be too small to allow the \ac{CMU} block to effectively capture the time dependencies in the section located between the encoder and decoder. 
After some preliminary tests, the number of convolutional filters has been set to 64 in the first layer of the encoder and in the subsequent blocks, while in the last layer it has been set equal to 16.
In this way, the dimensionality of the input vector goes from $I \in \mathbb{R}^{4 \times 16 \times 8 \times 2}$ to $O \in \mathbb{R}^{4 \times 4 \times 2 \times 16}$ as the encoder output. Symmetrically, the convolutions within the decoder use 64 filters, except for the final layer which uses only 2 filters to generate the prediction of the Inflow and Outflow channels. 
The parameters corresponding to the dimensionality of the convolution kernel (kernel size equal to 3, batch size equal to 16 and learning rate equal to 0.0001. The number of epochs is set to 150), to the batch size and to the learning rate have been optimized with the \textit{Bayesian optimization} technique. 

As for the models from the literature, they have been arranged and trained following carefully the parameter values and indications reported in the respective publications. 

\begin{table}[t]
\footnotesize
\centering
\caption{Results obtained for the Bike NYC data set}
\begin{tabular*}{\textwidth}{@{\extracolsep{\fill}}c|ccc}
\toprule
Model    &    RMSE &     MAPE &     APE \\
\midrule
HA       &       6.56 &       26.46 &  4.09E+05 \\
ST-ResNet &  5.01±0.07 &  21.97±0.26 &  3.40E+05±4.06E+03 \\
MST3D    &  4.98±0.05 &  22.03±0.47 &  3.41E+05±7.26E+03 \\
3D-CLoST  &  4.90±0.04 &  21.38±0.20 &  3.31E+05±3.12E+03 \\
PredCNN  &  4.81±0.04 &  21.38±0.24 &  3.31E+05±3.76E+03 \\
ST-3DNet  &  4.75±0.06 &  21.42±0.28 &  3.31E+05±4.36E+03 \\
STAR     &  4.73±0.05 &  20.97±0.13 &  3.24E+05±2.02E+03 \\
\hline
\modelname{}   &  \textbf{4.67±0.03} &  \textbf{20.85±0.15} &  \textbf{3.23E+05±2.31E+03} \\
\bottomrule
\end{tabular*}
\vspace*{2mm}
\label{tab:bikenyc}
\end{table}

As shown in \autoref{tab:bikenyc}, \modelname{} outperforms all other considered approaches for all evaluation metrics. In addition, the small standard deviation values are evidence of the robustness of the proposed approach.  Nonetheless, it is worth observing that all learning-based approaches return similar results. We believe this is mainly due to the reduced size of the data set that does not allow the models to be adequately trained. Moreover, the tessellation used in this case study (widely used in the literature), with a small grid of dimensions ($16\times 8$), tends to level off the metrics and hinder a more precise performance assessment.

\subsubsection{TaxiBJ}

As with the experiment discussed above, for the TaxiBJ case study the parameters of the models have been set according to the specifications given in the respective publications. In the case of \modelname{},  the hyperparameters are kept unchanged in the two experiments, except for the number $L$ of encoding and decoding blocks, which has been increased to 3 because the grid is larger ( $32\times 32$) in this experiment and more convolutional layers are needed to map the input tensor of the model. 
Also for this experiment, the kernel size, batch size, and learning rate parameters have been optimized with \textit{Bayesian optimization} and the best values found were 3, 16, and 0.0001 respectively. The number of epochs has been set at 150. Notice that these values are the same used in the BikeNYC experiment. 

\begin{table}[tb]
\footnotesize
\centering
\caption{Results obtained for the Taxi Beijing data set}
\begin{tabular*}{\textwidth}{@{\extracolsep{\fill}}c|ccc}
\toprule
Model    &    RMSE &     MAPE &     APE \\
\midrule
HA       &      40.93  &       30.96 &  6.77E+07 \\
ST-ResNet &  17.56±0.91 &  15.74±0.94 &  3.45E+07±2.05E+06 \\
MST3D    &  21.34±0.55 &  22.02±1.40 &  4.81E+07±3.03E+05 \\
3D-CLoST  &  17.10±0.23 &  16.22±0.20 &  3.55E+07±4.39E+05 \\
PredCNN  &  17.42±0.12 &  15.69±0.17 &  3.43E+07±3.76E+05 \\
ST-3DNet  &  17.29±0.42 &  15.64±0.52 &  3.43E+07±1.13E+06 \\
STAR     &  16.25±0.40 &  15.40±0.62 &  3.38E+07±1.36E+06 \\
\hline
\modelname{}   &  \textbf{15.61±0.11} &  \textbf{14.73±0.21} &  \textbf{3.22E+07±4.51E+05} \\
\bottomrule
\end{tabular*}
\vspace*{2mm}
\label{tab:taxibj}
\end{table}

As can be seen from \autoref{tab:taxibj}, \modelname outperforms all other methods, reducing by 6\%, 4.4\%, and 1.5\% RMSE, MAPE, and APE respectively, compared to the second-best approach. The difference in performance, in favor of the proposed model, in this experiment is more appreciable because the data set used for the training process is more significant but also because the number of regions is higher.  This last consideration highlights how the proposed model seems suitable to be applied in real-world scenarios, i.e., where high model accuracy and dense tessellation are required (i.e., the city is partitioned into a large number of small regions).


\subsubsection{TaxiNYC}

As mentioned earlier, the TaxiNYC case study was created specifically to be able to evaluate the behavior of the proposed model in a wider set of scenarios than the literature. Consequently, in order to make a fair comparison, it was necessary to search for the best configuration of hyperparameters not only for the \modelname{} model but also for all the other approaches considered.  The optimized parameters and the relative values used in the training phase are briefly summarized below for each model. Unreported configuration values have been set as for the BikeNYC case study since the two experiments share the same map size ($16\times 8$). The parameters for each model are as follows:
\begin{itemize}
\item \textbf{ST-ResNet*.} Optimized parameters: number of residual units, batch size and learning rate. Optimal values found: 2, 16 and 0.0001.
\item \textbf{MST3D.} Optimized parameters: batch size and learning rate. Optimal values found: 16 and 0.00034.
\item \textbf{PredCNN.} Optimized parameters: encoder length, decoder length, number of hidden units, batch size and learning rate. Optimal values found: 2, 3, 64, 16 and 0.0001.
\item \textbf{ST-3DNet.} Optimized parameters: number of residual units, batch size and learning rate. Best values found: 5, 16 and 0.00095.
\item \textbf{STAR*.} Optimized parameters: number of remaining units, batch size and learning rate. Optimal values found: 2, 16 and 0.0001.
\item \textbf{3D-CLoST.} Optimized parameters: number of \ac{LSTM} layers, number of hidden units in each LSTM layer, batch size, and learning rate. Optimal values found: 2, 500, 16, and 0.00076.
\item \textbf{\modelname.} Optimized parameters: kernel size, batch size, and learning rate. Optimal values found: 3, 64 and 0.00086.
\end{itemize}

It is worth noting that, preliminary experiments showed a convergence issue for the training phase of both STAR and ST-ResNet models. In particular, they were unable to converge for any combination of parameters. This behavior is due to the strong presence of \textit{outliers} and to the concentration of the relevant Inflow and Outflow values in a few central regions of the city. 
To overcome this issue, Batch Normalization layers have been inserted in the structure of the two models. In particular, Batch Normalization layers have been added after each convolution present in the residual units (a possibility that has already been foreseen in the original implementations) and after the terminal convolution of the networks (an option not considered in the source code provided by the original authors). 
For this reason, ST-ResNet and STAR are marked with an asterisk in the \autoref{tab:taxinyc}, which summarizes the experimental results.

\begin{table}[t]
\centering
\footnotesize
\caption{Results obtained for the Taxi New York data set}
\begin{tabular*}{\textwidth}{@{\extracolsep{\fill}}c|ccc}
\toprule
Model    &    RMSE &     MAPE &     APE \\
\midrule
HA       &     164.31 &       27.19 &  7.94E+05 \\
ST-ResNet* &  \textbf{35.87±0.60} &  22.52±3.43 &  6.57E+05±1.00E+05 \\
MST3D    &  48.91±1.98 &  23.98±1.30 &  6,98E+05±1.34E+04\\
3D-CLoST  &  48.17±3.16 &  22.18±1.05 &  6.48E+05±3.08E+04 \\
PredCNN  &  40.91±0.51 &  25.65±2.16 &  7.49E+05±6.32E+04 \\
ST-3DNet  &  41.62±3.44 &  25.75±6.11 &  7.52E+05±1.78E+05 \\
STAR*     &  36.44±0.88 &  25.36±5.24 &  7.41E+05±1.53E+05 \\
\hline
\modelname{}   &  36.22±0.72 &  \textbf{20.29±1.48} &  \textbf{5.93E+05±4.31E+04} \\
\bottomrule
\end{tabular*}
\vspace*{2mm}
\label{tab:taxinyc}
\end{table}

As it can be seen from \autoref{tab:taxinyc}, \modelname achieved excellent results in this experiment as well, ranked as the best model for two out of three evaluation metrics. In particular, as far as the RMSE is concerned, the performances obtained are very close to the best one (achieved by ST-ResNet*, which is considerably different from the original ST-ResNet), while the MAPE and APE values position it as the best model. 




\subsection{Ablation study}

In this section, an ablation study conducted on \modelname{} is presented in which variations in the input structure and in the network architecture are analyzed. 
The study, for reasons of space refers only to BikeNYC case study and does not involves the full combinatorics of all possible variants of the proposed model but aims to assess the impact on performance metrics of some parameters (namely, the number of input time points $n$) and specific architectural choices (viz., long skip connection, attention blocks, and external factors input branch), while maintaining all other conditions.  
More precisely, in what follows \modelname{} is compared against the 5 different variations described below:
\begin{itemize}
    \item \textbf{\modelname{}\_N3}. Same architecture as \modelname{}, but input volumes with 3 frames ($[X_{t-3}, X_{t-2}, X_{t-1}]$).
    \item \textbf{\modelname{}\_N5}. Same architecture as \modelname{}, but input volumes with 5 frames ($[X_{t-5}, X_{t-4}, X_{t-3}, X_{t-2}, X_{t-1}]$).
    \item \textbf{\modelname{}\_NoLSC}. \modelname{} by removing the \textit{long skip connection} between encoder and decoder.
    \item \textbf{\modelname{}\_NoAtt}. \modelname{} without the \textit{attention} blocks.
    \item \textbf{\modelname{}\_NoExt}. \modelname{} without the external factors.
\end{itemize}

Notice that the study does not consider the variations with $n=1$ and $n=2$ as such values would not allow the network to capture meaningful temporal patterns between traffic flows.

\begin{table}[t]
\footnotesize
\centering
\caption{Results obtained from ablation studies}
\begin{tabular*}{\textwidth}{@{\extracolsep{\fill}}c|ccc}
\toprule
Model    &    RMSE &     MAPE &     APE \\
\midrule
\modelname{}\_N3  & 4.75±0.04  &  21.18±0.18 & 3.28E+05±2.73E+03\\
\modelname{}\_N5    & 4.74±0.03 & 21.03±0.22 &  3.26E+05±3.43E+03 \\
\modelname{}\_NoLSC    & 4.84±0.04 & 21.53±0.24 & 3.33E+05±3.71E+03 \\
\modelname{}\_NoAtt &  4.78±0.04 & 20.95±0.27 &  3.25E+05±4.20E+03 \\
\modelname{}\_NoExt &  4.76±0.04 & 20.99±0.29 &  3.26E+05±4.55E+03 \\
\hline
\modelname{}       &  \textbf{4.67±0.03} & \textbf{20.85±0.15} & \textbf{3.23E+05±2.31E+03} \\
\bottomrule
\end{tabular*}
\vspace*{2mm}
\label{tab:ablation}
\end{table}

\autoref{tab:ablation} reports the results of the ablation study conducted. Each data point in the table has been obtained performing 10 times the training procedure for each model variation changing the random seed, and evaluating the resulting network on the test set. The mean and standard deviation are reported. 

The results show that regarding the time horizon, for the BikeNYC case study, $n=4$ allows the model to obtain better results. This means that, considering the particular setup, for the city of New York 4 hours of data allow to predict more accurately the dynamics of bicycle mobility whereas considering a greater amount of information ($n=5$) would reduce the accuracy of the network. It is plausible to believe that considering a larger number of temporal instants would lead the network to grow in the number of parameters to be trained and thus require a larger amount of data to identify possible longer-term patterns. 


From the architectural point of view, the two components attention block and long skip connection, confirm their importance in improving the performance of the proposed model, accounting for a 2.36\% and 3.64\% increase in RMSE, respectively. In particular, as regards the attention block, not only \modelname{} reaches lower average error values, but also the standard deviation is reduced, proving that the  attention blocks are effective in helping the network single out the most meaningful information and in making the training process more stable.
Finally, the experiment shows also a strong impact of the long skip connection mechanism, which, as illustrated in \autoref{sec: decoder}, connects the encoder to the decoder to convey fine-grained details through the network.

\subsection{Number of trainable parameters and FLOPs}

A a brief analysis of the number of trainable parameters and the computational complexity (measure in number of FLOPs) of each model for the different case studies is reported in this section. 
For what concerns the number of trainable parameters, as shown in \autoref{tab:parameters}, \modelname{} has a generally low number compared to other models as only STAR features lesser parameters to train. 
Such a reduced number of parameters is due to the fact that the dimensionality of the input is reduced by the encoder downsampling mechanism.
The model with the highest number of parameters is 3D-CLoST, which uses both 3D convolutions and \ac{LSTM}.

Finally, \autoref{tab:flops} provide the computational complexity of each model in terms floating point operations (FLOPs) as in \cite{8506339} for each case study. 
As can be seen from the results obtained, the model with the higher computational complexity is PredCNN, which is based on CMUs. While, 3DCLoST is the model with the shortest forward and backward times. \modelname{}, instead, has a middle-range computational complexity compared to the other models, despite its autoencoder structure, the use of attention blocks, and CMUs. This occurs because although the number of network parameters is small, the network employs high complexity operators. However, the training and execution times of \modelname{} are compatible with its applicability in full-scale real-world scenarios.

\begin{table}[t]
\footnotesize
\centering
\caption{Number of trainable parameters}
\begin{tabular*}{\textwidth}{@{\extracolsep{\fill}}c|rrr}
\toprule
{Model} & {BikeNYC} & {TaxiNYC} & {TaxiBJ} \\ 
\midrule
{ST-ResNet} &  906,272 & 458,304 & {2,696,992} \\
{MST3D} & 668,218 & 668,378 & {8,674,370} \\
{3D-CLoST} & 13,099,090 & 19,477,648 & {72,046,714} \\
{PredCNN} & 3,967,906 & 3,967,906 & {4,827,842} \\
{ST-3DNET} & 540,696 & 617,586 & {903,242} \\
{STAR} & 161,052 & 310,076 & {476,388} \\ \hline
{\modelname{}} & 582,673 & 582,673 & {765,497} \\ \bottomrule
\end{tabular*}
\label{tab:parameters}
\end{table}

\begin{table}[t]
\footnotesize
\centering
\caption{Computational complexity (in number of FLOPs)}
\begin{tabular*}{\textwidth}{@{\extracolsep{\fill}}c|rrr}
\toprule
Model & 
{BikeNYC} & {TaxiNYC} & {TaxiBJ} \\ 
\midrule
{ST-ResNet} & 230,849,450 & 115,735,786 & {5,459,663,018} \\
{MST3D} & 33,042,250 & 33,042,570 & {272,483,226} \\
{3D-CLoST} & 29,613,094 & 9,601,920 & {338,148,804} \\
{PredCNN} & 1,015,468,288 & 1,015,468,288 & {9,883,813,888} \\
{ST-3DNET} & 171,242,496 & 190,130,922 & {1,823,295,898} \\
{STAR} & 40,449,706 & 78,231,530 & {928,100,922} \\ \hline
{\modelname{}} & 130,047,738 & 130,047,738 & {1,067,063,882} \\ 
\bottomrule
\end{tabular*}
\vspace*{2mm}
\label{tab:flops}
\end{table}

\section{Conclusions}\label{sec:conclusions}
Predicting vehicular flow is one of the central topics in the domain of intelligent mobility. It is a challenging task, influenced by several complex factors, such as spatio-temporal dependencies and external factors.
In this study, we have developed a new deep learning architecture, based on convolutions and \ac{CMU} to forecast the Inflow and Outflow in each region of the smart city.
A comprehensive experimental campaign has been conducted on three different real-world case studies. The results showed that \modelname consistently outperforms state-of-the-art models in predicting dynamics in all the experiments conducted on the three performance metrics considered.  This work also reports and analyzes the results from an ablation study and complexity analysis. 

For future developments, the possible integration of other external factors, such as the territorial characteristics of each geographical area, should be tested. Moreover, it would be appropriate to increase the granularity of the city tessellation, as well as conduct transfer learning experiments to study the applicability of the proposed model to scenarios with a reduced amount of data available.



\bibliographystyle{elsarticle-num} 
\bibliography{IDBM}


\end{document}